\documentclass[10pt,twocolumn,letterpaper]{article}

\usepackage{cvpr}
\usepackage{times}
\usepackage{epsfig}
\usepackage{graphicx}
\usepackage{amsmath}
\usepackage{amssymb}

\usepackage{helvet}
\usepackage{courier}
\usepackage{graphicx}
\usepackage{bm}
\usepackage{color}
\usepackage{bbm}
\usepackage{epstopdf}
\usepackage{caption}
\usepackage{enumitem}
\usepackage{calc}
\usepackage{multirow}
\usepackage{xspace}
\usepackage{booktabs}

\newcommand{\ve}[1]{{\mathbf #1}} % for displaying a vector or matrix
\newcommand{\hua}[1]{{\mathcal #1}}

\makeatletter
\DeclareRobustCommand\onedot{\futurelet\@let@token\@onedot}
\def\@onedot{\ifx\@let@token.\else.\null\fi\xspace}
\def\eg{\emph{e.g}\onedot} 
\def\ie{\emph{i.e}\onedot} 
 
\def\etc{\emph{etc}\onedot} 
\def\wrt{w.r.t\onedot} 
\def\etal{\emph{et al}\onedot}

\graphicspath{{./fig/}{}}

\usepackage[pagebackref=true,breaklinks=true,letterpaper=true,colorlinks,bookmarks=false]{hyperref}

\cvprfinalcopy % *** Uncomment this line for the final submission

\def\cvprPaperID{3111} % *** Enter the CVPR Paper ID here

\ifcvprfinal\fi
\begin{document}

%%%%%%%%% TITLE
\title{Joint Multi-Person Pose Estimation and Semantic Part Segmentation}
\author{
Fangting Xia$^{1}$~~~~~~~~~~~~~~~~~~~~~Peng Wang$^{1}$~~~~~~~~~~~~~~~~~~Xianjie Chen$^{1}$~~~~~~~~~~~Alan Yuille$^{2}$~~~\\
{\tt\small sukixia@gmail.com~~~pengwangpku2012@gmail.com~~~cxj@ucla.edu~~~alan.yuille@jhu.edu}\\
\\
$^{1}$University of California, Los Angeles~~~~~~~~~~~~~~~~$^{2}$Johns Hopkins University\\
~~~~~~~~~Los Angeles, CA 90095~~~~~~~~~~~~~~~~~~~~~~~~~~~~~~~~Baltimore, MD 21218
}
%\author{Fangting Xia\\
%University of California, Los Angeles\\
%Los Angeles, CA 90095\\
%{\tt\small sukixia@gmail.com}
%\and
%Peng Wang\\
%University of California, Los Angeles\\
%Los Angeles, CA 90095\\
%{\tt\small pengwangpku2012@gmail.com}
%\and
%Alan Yuille\\
%Johns Hopkins University\\
%Baltimore, MD 21218\\
%{\tt\small alan.yuille@jhu.edu}
%}

\maketitle
%\thispagestyle{empty}

%%%%%%%%% ABSTRACT
\begin{abstract}
Human pose estimation and semantic part segmentation are two complementary tasks in computer vision. In this paper, we propose to solve the two tasks jointly for natural multi-person images, in which the estimated pose provides object-level shape prior to regularize part segments while the part-level segments constrain the variation of pose locations. Specifically, we first train two fully convolutional neural networks (FCNs), namely Pose FCN and Part FCN, to provide initial estimation of pose joint potential and semantic part potential. Then, to refine pose joint location, the two types of potentials are fused with a fully-connected conditional random field (FCRF), where a novel segment-joint smoothness term is used to encourage semantic and spatial consistency between parts and joints. To refine part segments, the refined pose and the original part potential are integrated through a Part FCN, where the skeleton feature from pose serves as additional regularization cues for part segments. Finally, to reduce the complexity of the FCRF, we induce human detection boxes and infer the graph inside each box, making the inference forty times faster.

Since there's no dataset that contains both part segments and pose labels, we extend the PASCAL VOC part dataset~\cite{chen2014detect} with human pose joints\footnote{\url{https://sukixia.github.io/paper.html}} and perform extensive experiments to compare our method against several most recent strategies. We show that our algorithm surpasses competing methods by 10.6$\%$ in pose estimation with much faster speed and by 1.5$\%$ in semantic part segmentation.
\end{abstract}

\section{Introduction}
% Definition of the two tasks. There're many applications from the two tasks. The two tasks are challenging in multi-person scenes. However, they can help each other.
Human pose estimation (\ie predicting the position of joints for each human instance) and semantic part segmentation (\ie decomposing humans into semantic part regions) are two crucial and correlated tasks in analysing humans from images. They provide richer representations for many dependent tasks, \eg fine-grained recognition~\cite{branson2014bird,zhang2014part,krause2015unreasonable}, action recognition~\cite{wang2012discriminative,wang2013approach}, image/video retrieval~\cite{yamaguchi2015retrieving,jones2013content}, person-identification~\cite{ma2011human} and video surveillance~\cite{lu2014application}.

% Intuition and big picture
\begin{figure}[!tb]
\begin{center}
   \includegraphics[width=1.0\linewidth]{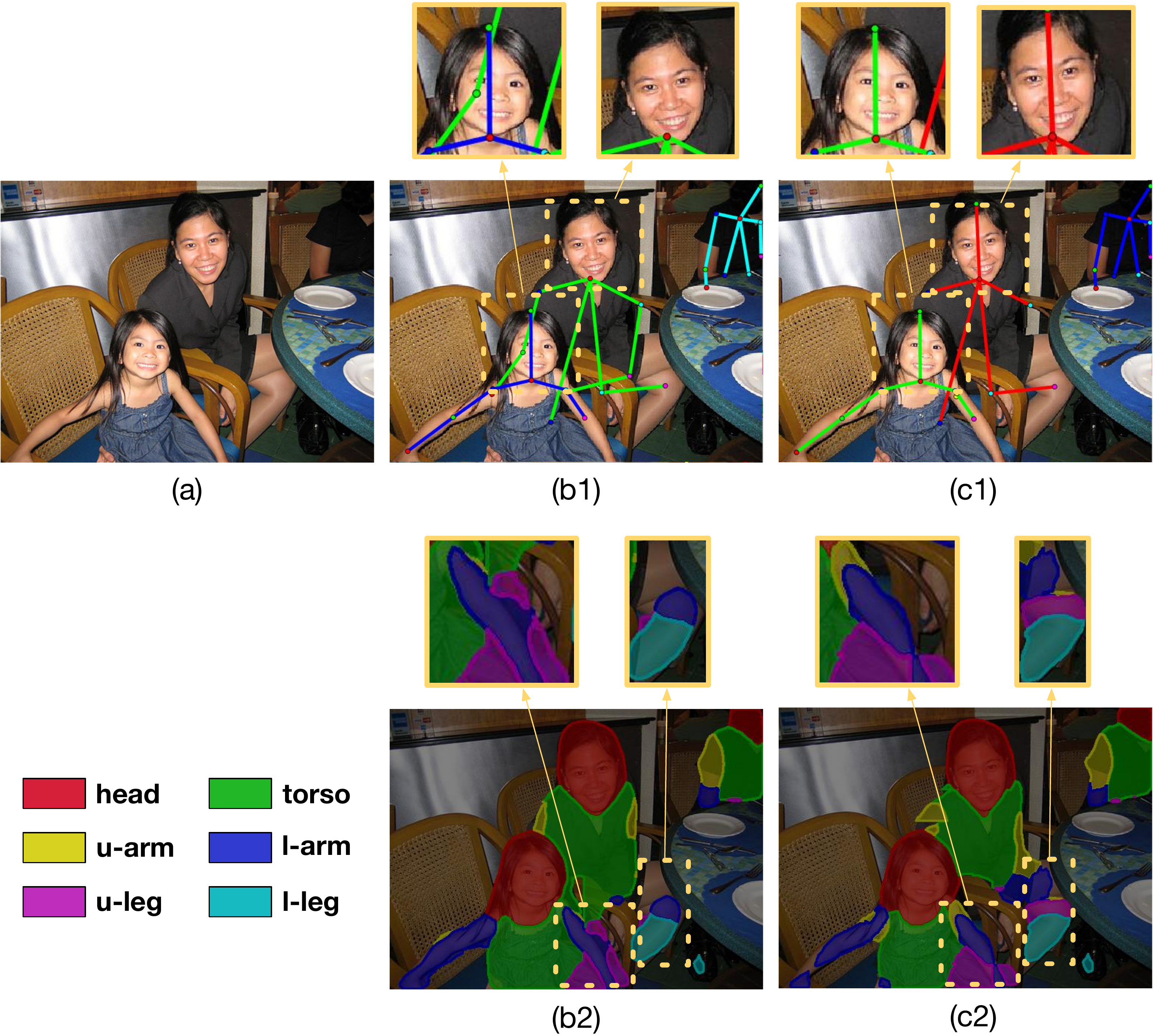}
\end{center}
\caption{Joint human pose estimation and semantic part segmentation improve both tasks. (a) input image. (b) pose estimation and semantic part segmentation results before joint inference. (c) pose estimation and semantic part segmentation results after joint inference. Note that comparing (b1) and (c1), our result recovers the missing forehead joint and corrects the location error of right elbow and right wrist for the woman on the right. Comparing (b2) and (c2), our result gives more accurate details of lower arms and upper legs than (b2) for both people.}
\label{fig:intuition}
\end{figure}

Recently, dramatic progress has been made on pose estimation~\cite{chen2014articulated,Chen_CVPR15,xia2016pose,newell2016stacked} and human part segmentation~\cite{chen2014semantic,wang2015joint,xia2015zoom,liang2016semantic} with the advent of powerful convolutional neural networks (CNN)~\cite{lecun2010convolutional} and the availability of pose/segment annotations on large-scale datasets~\cite{everingham2014pascal,chen2014detect,lin2014microsoft}.
However, the two tasks are mostly solved independently without considering their correlations. As shown in the middle column in Fig.~\ref{fig:intuition}, for pose estimation, by designing loss \wrt the joints solely, it may omit the knowledge of dense pixel-wise part appearance coherence, yielding joints located outside of human instance or misleading joints when two people are close to each other. On the other hand, for part segmentation, through training that only respects pixel-wise part labels, it lacks proper overall human shape regularization, yielding missing/errorneous predictions when appearance cues are weak or missing.
%But there remains serious problems when dealing with images containing multiple people, who may partially occlude each other, and particularly if the scales and poses of the people vary. These are the situations which we address in this paper.

%Both tasks face challenges if addressed individually. 
In fact, the two tasks are complementary, and solving them jointly can reduce the learning difficulty in addressing each of them individually.
%Pose estimation can yield poor localization of joints (\eg outside the person) and may fail to detect some joints, such as the knee joint if a person is wearing a long dress. Semantic part segmentation can fail to distinguish ambiguous regions in semantic part segmentation when the person is in non-typical pose. 
%But it is easier to handle these difficult cases if we allow the two tasks to cooperate.
As shown in the right column Fig.~\ref{fig:intuition}, by handling the two tasks jointly, the ambiguity in pose estimation (\eg out of instance region) can be corrected by considering semantic part segments, while the estimated pose skeleton provides object-level context and regularity to help part segments align with human instances, \eg over the details of arms and legs where appearance cues are missing.

\begin{figure*}[!tb]
\begin{center}
   \includegraphics[width=0.95\textwidth]{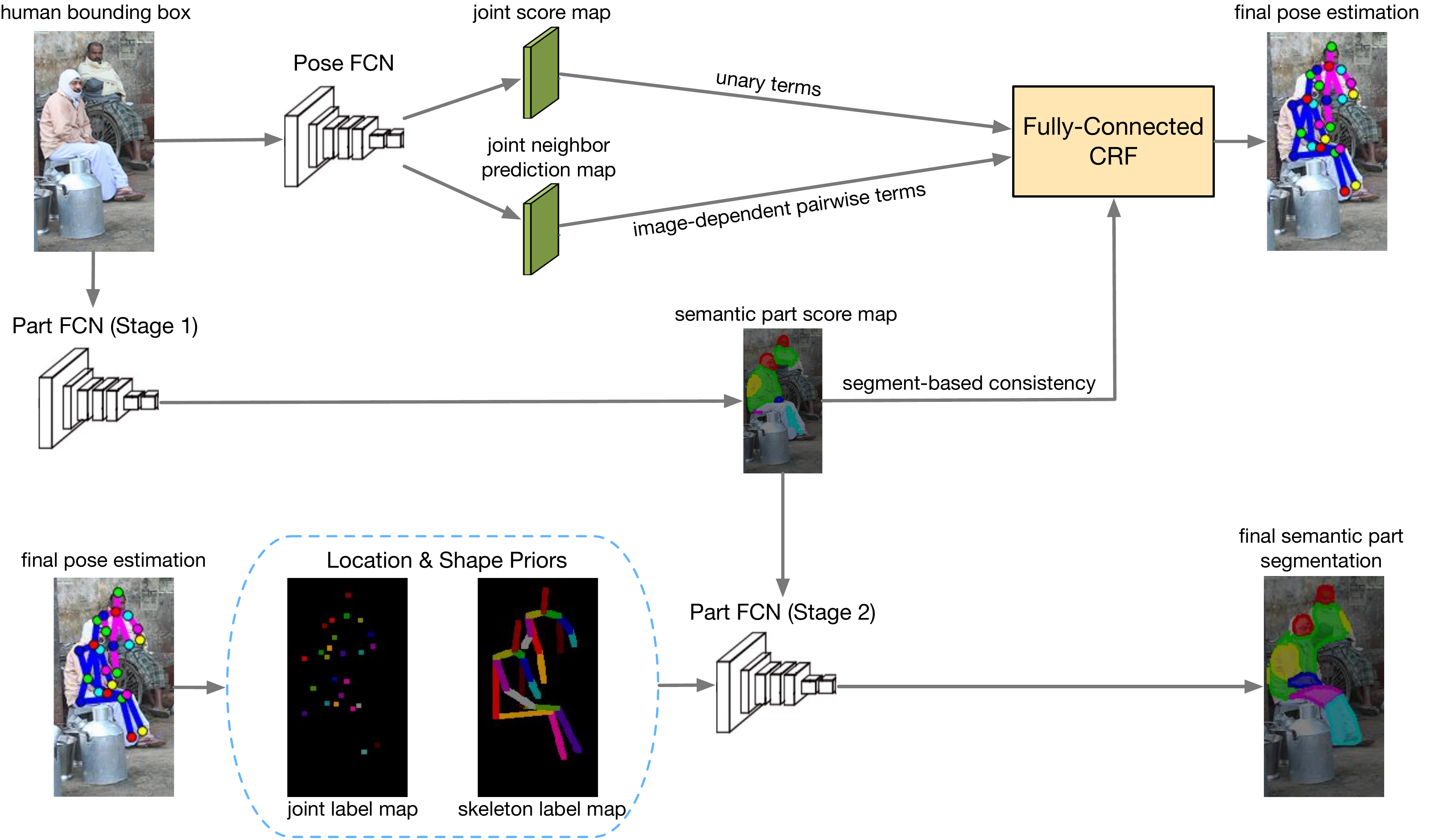}
\end{center}
\caption{The framework of our approach for joint pose estimation and part segmentation. Initial joint scores and part segment scores are fused to yield better pose estimation results, and then the estimated poses are used to refine part segmentation.}
\label{fig:framework}
%\vspace{-10pt}
\end{figure*}

Specifically, we illustrate our framework in Fig.~\ref{fig:framework}. Firstly, given an image that contains multiple people, we train two FCNs: Pose FCN and Part FCN. Similar to~\cite{insafutdinov2016deepercut}, the Pose FCN outputs the pixel-wise joint score map, \ie the potential of joints at each pixel (how likely a type of joint is located at certain pixel), and also outputs the joint neighbour score map, \ie the potential of the location likelihood of neighboring joints for each joint type. The Part FCN produces the part score map for each semantic part type. Secondly, the three types of information are fused through a FCRF to refine the human joint locations, where a novel smoothness term on both part segments and joint proposals (generated from the initially estimated pixel-wise joint score map) are applied to encourage the consistency between segments and joints. Thirdly, the refined pose joints are re-organized into pose features that encode overall shape information, and are fed into a second-stage Part FCN as an additional input besides the initial part score map, yielding better segmentation results. To reduce the complexity of the FCRF, rather than infer over the full image as~\cite{insafutdinov2016deepercut}, we adopt a human detector~\cite{ren2015faster} to first get the bounding box for each human instance and resize each instance region in a similar way to~\cite{xia2015zoom}. Our whole inference procedure is then performed within each resized region. 

%In our pose estimation model, we propose to: (1) model consistency with semantic part segmentation results by incorporating segment-based terms into a fully-connected CRF while penalizing joint candidate pairs whose locations do not agree with their corresponding semantic parts; (2) employ an ``auto-zoom'' strategy to first detect human bounding boxes and then perform multi-person pose estimation within each properly resized bounding box. In our semantic part segmentation model, we use top-down pose cues, and also use the ``auto-zoom'' idea. More precisely, for each resized human bounding box, we infer feature maps that capture joints and skeleton information from good pose estimates, and then feed these feature maps together with the original image to a dedicated fully convolutional network (FCN) to refine the part segmentation results. Our model is able to correct severe pose errors and gives clearer details of small, but important, parts like arms and legs.

Last but not the least, in order to train and evaluate our method, we augment the challenging PASCAL-Person-Part dataset~\cite{chen2014detect} with 14 human pose joint locations through manual labeling and make the annotations public. This dataset includes 3533 images that contain large variation of human poses, scales and occlusion. We evaluate our method over this dataset, and show that our approach outperforms the most recent competing methods for both tasks. In particular, our method is more effective and much faster (8 seconds versus 4 minutes) than DeeperCut~\cite{insafutdinov2016deepercut} which is arguably the most effective algorithm for multi-person pose estimation.

In summary, the contributions of this paper lay in three folds: (1) to our best knowledge, we are the first to explore and demonstrate the complementary property of multi-person pose estimation and part segmentation with deep learned potentials; (2) by combining detection boxes in the pipeline, we reduce the complexity of FCRF inference over the full image, yielding better efficiency; (3) we extend the well labelled PASCAL-Person-Part dataset with human joints and demonstrate the effectiveness of our approach.

% framework

\section{Related Works}
% Pose estimation: traditional methods and currently prevailing methods.
\paragraph{Pose estimation.} Traditional approaches use graphical models to combine spatial constraints with local observations of joints, based on low-level features~\cite{felzenszwalb2005pictorial,yang2011articulated}.
With the growing popularity of deep learning, recent methods rely on strong joint detectors trained by DCNNs~\cite{chen2014articulated,tompson2015efficient}, and often use a simple graphical model (\eg tree model, And-Or graph) to select and assemble joints into a valid pose configuration. These recent methods perform much better than traditional ones, but the localization of joints is still inaccurate (\eg sometimes outside the human body) and they still struggle when there are multiple people overlapping each other. Other approaches discard graphical models by modeling the spatial dependencies of joints within DCNNs~\cite{toshev2014deeppose,carreira2015human,chu2016structured}. These approaches perform well on relatively simple datasets, but their ability to handle large pose variations in natural multi-person datasets is limited. A very recent work, Deeper-Cut~\cite{insafutdinov2016deepercut}, addresses the multi-person issue explicitly, using integer linear programming to cluster joint candidates into multiple human instances and assign joint types to each joint candidate.
Deeper-Cut handles multi-person overlapping well, but is very time-consuming (4 minutes per image) and its performance on datasets with large scale variation is not fully satisfactory. Our method improves in these aspects by introducing a segment-joint consistency term that yields better localization of flexible joints such as wrists and ankles, and an effective scale handling strategy (using detected boxes and smart box rescaling) that can deal with humans of different sizes.  

% Semantic part segmentation: traditional methods (proposals by graph cuts + complicated graphical models) not suited for complicated multi-person datasets, and time-consuming;
% With the advent of deep-learning, FCN-type methods provide easy end-to-end solutions. However, there're still much space for improvement when the dataset has large variation in scale, pose, multi-person occlusion.
% Our method: use top-down pose information to relieve disambiguation and severe pose errors.
%\vspace{-10pt}
\paragraph{Semantic part segmentation.} Previous approaches either use graphical models to select and assemble region proposals~\cite{xia2016pose}, or use fully convolutional neural networks (FCNs)~\cite{long2014fully} to directly produce pixel-wise part labels. Traditional graphical models~\cite{yamaguchi2012parsing,dong2013deformable} find it difficult to handle the large variability of pose and occlusion in natural images. FCN-type approaches~\cite{chen2014semantic,wang2015joint}, though simple and fast, give coarse part details due to FCN's inherent invariance property, and can have local confusion errors (\eg labeling arms as legs, labeling background regions as arms, \etc) if the person is in a non-typical pose, or when there are some other object/person nearby with similar appearance. Two recent works improve on FCN-type approaches by paying attention to the large scale variation in natural images. Chen \etal learn pixel-wise weights through an attention model~\cite{chen2015attention} to combine the part segmentation results of three fixed scales. Xia \etal build an hierarchical model that adapts to object scales and part scales using ``auto-zoom''~\cite{xia2015zoom}. We treat these two methods as our baselines, and demonstrate the advantages of our part segmentation approach. Most recently, researchers design and adopt more powerful network architectures such as Graph Long Short-Term Memory (LSTM)~\cite{liang2016semantic} and DeepLab with Deep Residual Net~\cite{chen2016deeplab}, greatly improving the performance. We prove that our method is complementary and can be added to these networks to further improve the performance.

%\vspace{-10pt}
\paragraph{Joint pose estimation and part segmentation.} Yamaguchi \etal perform pose estimation and semantic part segmentation sequentially for clothes parsing, using a CRF with low-level features~\cite{yamaguchi2012parsing}. Ladicky \etal combine the two tasks in one principled formulation, using also low-level features~\cite{ladicky2013human}. Dong \etal combine the two tasks with a manually designed And-Or graph~\cite{dong2014towards}. These methods demonstrate the complementary properties of the two tasks on relatively simple datasets, but they cannot deal with images with large pose variations or multi-person overlapping, mainly due to the less powerful features they use or the poor quality of their part region proposals. In contrast, our model combines FCNs with graphical models, greatly boosting the representation power of models to handle large pose variation. We also introduce novel part segment consistency terms for pose estimation and novel pose consistency terms for part segmentation, further improving the performance.

\section{Our Approach}
Given an image $\ve{I}$ with size h$\times$w, our task is to output a pixel-wise part segmentation map $\ve{L}_{s}$, and a list of scored pose configurations $\hua{C}_{p}=\{(\ve{c}_i, s_i)|i=1,2,\dots, k_i\}$, where $\ve{c}_i$ is the location of all 14 pose joint types for the person and $s_i$ is the score of this pose configuration.

As illustrated in Fig.~\ref{fig:framework}, for each human detection box, we first use Pose FCN and Part FCN to give initial estimation of pose location and part segmentation. Then a FCRF is used to refine pose estimation and a second-stage Part FCN is adopted for part refinement.
Specifically, we first extract human bounding boxes with Faster R-CNN~\cite{ren2015faster}, and resize the image region within each detection box following~\cite{xia2015zoom} so that small people are enlarged and extra large people are shrunk to a fixed size. The resized box regions serve as input to Pose FCN and Part FCN.
Pose FCN adopts the network architecture of ResNet-101 proposed in~\cite{he2015deep}, while for Part FCN we use DeepLab-LargeFOV~\cite{chen2014semantic}.

Pose FCN outputs two feature maps: (1) the pixel-wise joint score map $\ve{P}_j$, which is a matrix with shape h$\times$w$\times$ 14 representing the probability of each joint type locating at each pixel. (2) the pixel-wise joint neighbor score map $\ve{P}_n$, which is a h$\times$w$\times$364 matrix representing the probability of expected neighbor location for each joint. Here, the dimension of 364 is obtained by 14$\times$13$\times$2, which means for each joint the we estimate the other 13 joint locations using the offset $(\delta x, \delta y)$. Following the definition of parts in~\cite{chen2014semantic}, Part FCN outputs a part score map $\ve{P}_s$ including 7 classes: 6 part labels and 1 background label.

Given the three score maps, we design a novel segment-joint smoothness term for our FCRF to obtain refined pose estimation results (detailed in Sec.~\ref{subsec:humanpose}). To obtain better part segmentation results, we further design a second-stage Part FCN, which takes joint input of first-stage part scores and derived feature maps from refined poses (detailed in Sec.~\ref{subsec:part}). Finally, the estimated poses from each bounding box are merged through a Non-Maximum Suppression (NMS) strategy detailed in Sec.~\ref{sec:exp_details}. For part segmentation, we merge the segment score map from different boxes using score averaging similar to~\cite{xia2015zoom}.

\subsection{Human Pose Estimation}
\label{subsec:humanpose}
% joint proposals from the joint score map; CRF with unary term and pairwise terms.
In this section, we explain how we unify the three score maps (\ie $\ve{P}_j$, $\ve{P}_n$ and $\ve{P}_s$) to estimate poses in each human detection box.

Following DeeperCut~\cite{insafutdinov2016deepercut}, we adopt a FCRF to obtain robust context for assembling the proposed joints into human instances.
To reduce the complexity of the FCRF, rather than consider all the pixels, we generate 6 candidate locations for each joint from the joint score map $\ve{P}_j$ by non-maximum suppression (NMS). Formally, the FCRF for the graph is formulated as $\hua{G} = \{\hua{V}, \hua{E}\}$, where the node set $\hua{V} = \{c_1,c_2,\dots,c_n\}$ represents all the candidate locations of joints and the edge set $\hua{E} = \{(c_i,c_j) | i=1,2,\dots,n, j=1,2,\dots,n, i<j\}$ is the edges connecting all the locations.
The label to predict for each node is its joint type $l_{c_i}\in\{0, \cdots, K\}$, where $K=14$ is the number of joint types and type $0$ represents that the node belongs to background and is not selected. Besides, we also predict whether two nodes belong to the same person, \ie $l_{c_i,c_j}\in\{0, 1\}$, where $1$ indicates the two nodes are for the same person. Let $\hua{L}=\{l_{c_i} | c_i \in \hua{V}\} \cup \{l_{c_i,c_j} | (c_i,c_j) \in \hua{E}\}$. The target we want to optimize is:
\begin{align}
\label{eqn:crf}
\min_\hua{L} \sum_{c_i \in \hua{V}} \psi_i(l_{c_i}) + \sum_{(c_i,c_j) \in \hua{E}} \psi_{i,j}(l_{c_i}, l_{c_j}, l_{c_i,c_j})
\end{align}
where the unary term is defined as $\psi_i = \log \frac{1-\ve{P}_j(l_{c_i})}{\ve{P}_j(l_{c_i})}$, which is a log-likelihood at location $c_i$ based on the Pose-CNN output, the joint score map $\ve{P}_j$.

In contrast, the pairwise term is determined by both the joint neighbor score map $\ve{P}_n$ and the segmentation score map $\ve{P}_s$. Formally,
\begin{align}
&\psi_{i,j}=l_{c_i,c_j}\log \frac{1-\ve{P}_{i,j}(l_{c_i}, l_{c_j} | \ve{P}_n, \ve{P}_s)}{\ve{P}_{i,j}(l_{c_i}, l_{c_j} | \ve{P}_n, \ve{P}_s)}
\end{align}
where $\ve{P}_{i,j}(l_{c_i}, l_{c_j})=\frac{1}{1+\exp(-\bm{\omega}\cdot \ve{f}(c_i, c_j, l_{c_i}, l_{c_j}))}$, obtained from logistic regression results \wrt a combined feature vector $\ve{f}$ from $\ve{f}(\ve{P}_n)$ and $\ve{f}(\ve{P}_s)$, in which we omit $c_i, c_j, l_{c_i}, l_{c_j}$ for simplicity.

% traditional features and segment-based features
The feature vector $\ve{f}(\ve{P}_n)$ encodes information to help decide whether the two proposals belong to the same person. We borrow the idea proposed in~\cite{insafutdinov2016deepercut}, and here we explain how the feature is extracted for paper completeness. Given the location of two joint proposals $c_i, c_j$ and their corresponding label $l_{c_i}, l_{c_j}$, we first derive a direct vector from $c_i$ to $c_j$, denoted as $\ve{v}_{i,j}$. In addition, given $c_i, l_{c_i}, l_{c_j}$, based on the joint neighbor offset score map $\ve{P}_n$, we may find an estimated location of $l_{c_j}$ respecting $c_i$ though computing $c_j' = c_i + (\delta x, \delta y)_{i, j}$. We denote the direct vector from $c_i$ to the estimated location as $\ve{v}'_{i,j}$. Similar vectors $\ve{v}_{j,i}, \ve{v}'_{j,i}$ can be extracted in the same way. Feature $\ve{f}(\ve{P}_n)=[$ $|\ve{v}_{j,i}-\ve{v}'_{j,i}|$,  $|\ve{v}_{i,j}-\ve{v}'_{i,j}|$, $<\ve{v}_{j,i},\ve{v}'_{j,i}>$, $<\ve{v}_{i,j},\ve{v}'_{i,j}>$ $]$, in which $|. - .|$ is the euclidean distance between two vectors and $<.$ $,$ $.>$ is the angle between two vectors.
%$\bm{f}$ uses a 6-d feature to describe the angle and distance between two pairs of vectors ($\bm{g_i g_j}$ vs. $\bm{g_i \tilde{g_j}}$, and $\bm{g_j g_i}$ vs. $\bm{g_j \tilde{g_i}}$).

The feature vector $\ve{f}(\ve{P}_s)$ considers the correlation between joints and segments. Intuitively, joints are the connection points of parts. If two joints are neighboring joints, using forehead and neck as an example, the head joint should be located inside the head segment region and near the head segment boundary while the neck joint should be located in either head or body region and near the common boundary of body and head. Moreover, the connected line between forehead joint and neck joint should fall inside the head region. These segment-based heuristic cues provide strong constrains for the location of joints. We design $\ve{f}(\ve{P}_s)$ \wrt this idea. Formally, each joint type is associated with one or two semantic parts and each neighbouring joint type pair is associated with one semantic part type.

Based on the part segmentation label map $\ve{L}_s$ inferred from $\ve{P}_s$, here we introduce the feature $\ve{f}(\ve{P}_s)$ using the example of forehead and neck. For details, please see the supplementary material.
Suppose $l_{c_i}$= forehead and $l_{c_j}$= neck, then our feature from segment includes 4 components: (1) a 2-d binary feature, with the first dimension indicating whether $c_i$ is inside the head region, and the second dimension indicating whether it is around the boundary of the head region;  (2) a 4-d binary feature, with the first 2-d feature indicating $c_j$ \wrt the head region same as (1), and the rest 2-d feature indicating $c_j$ \wrt the torso region respectively; (3) a 1-d feature indicating the proportion of pixels on the line segment between $c_i$ and $c_j$ that fall inside the head region; (4) a 1-d feature indicating the intersect-over-union (IOU) between an oriented rectangle computed from $c_i$ and $c_j$ (with aspect ratio = 2.5:1) and the head region. We only extract the full feature for neighboring joints. For the joints locating far away like head and feet, we drop the third and the fourth components of the feature and set them to be 0. We validate the parameters for aspect ratio through a mean human shape following~\cite{sigal2006predicting}.

%In this case, the segment-based consistency features are made up of: (1) binary features indicating the location of $g_i$ \wrt the head region: inside the region, around the boundary, or far away from the region; (2) binary features indicating the location of $g_j$ \wrt the head region and the torso region respectively; (3) the proportion of pixels on the line $\bm{g_i g_j}$ that fall inside the head region.

% inference and scoring
Based on the unary and pairwise terms described above, the FCRF infers the best labels $\hua{L}$ for the generated joint proposals $c_1,c_2,\dots,c_n$, selecting and assembling them into a list of pose configurations. We adapt the inference algorithm introduced in~\cite{insafutdinov2016deepercut}, transforming the FCRF into an integer linear programming (ILP) problem with additional constraints from $\hua{L}$. For each detection box, the inference algorithm gives the labels $\hua{L}$ for joint proposals within 1 sec. and we can acquire a list of pose configurations based on $\hua{L}$, with pose score equal to the sum of unary scores for all visible joints. For each detection box, we choose only one pose configuration whose center is closest to the detection box center, and add that pose configuration to our final pose estimation result. We also experiment with the strategy of extracting multiple pose configurations from each detection box since there might be multiple people in the detection box, but find this strategy doesn't improve the results.

\subsection{Semantic Part Segmentation}
\label{subsec:part}
% location and shape priors and the refinement network
We train a part segmentation model (the second-stage Part FCN) to segment an image into semantic parts with estimated high-quality pose configurations $\hua{C}_{p}$. We define two pose feature maps from $\hua{C}_{p}$: a joint label map and a skeleton label map, and use them as inputs to the second-stage Part FCN in addition to the original part score map. For the joint label map, we draw a circle with radius 3 at each joint location in $\hua{C}_{p}$. For the skeleton label map, we draw a stick with width 7 between neighbouring joints in $\hua{C}_{p}$. Fig.~\ref{fig:framework} illustrates the two simple and intuitive feature maps.

The second-stage Part FCN is much lighter than the first-stage Part FCN since we already have the part score map $\ve{P}_s$ predicted. We concatenate the 2 dimension feature map from estimated poses with the original part score map, yielding a 7 + 2 dimension inputs, and stacked 3 additional convolutional layers with kernel size as 7, kernel dimension as 128 and Relu as activation function. Our final part segmentation is then derived using the argmax value from the output part score map.

To learn all the parameters, we adopt a stage-wise strategy, \ie first learn Pose FCN and the first-stage Part FCN, then the FCRF, and finally the second-stage Part FCN, which roughly take 3 days to train. For inference, our framework takes roughly 6s per-image. It is possible for us to do learning and inference iteratively. However, we found it's practically inefficient and the performance improvement is marginal. Thus, we only do the refinement once.

\section{Experiments}\label{sec:exp}
\subsection{Implementation Details}\label{sec:exp_details}
\paragraph{Data.} We perform extensive experiments on our manually labeled dataset, PASCAL-Person-Part~\cite{chen2014detect}, which provides joint and part segment annotations for PASCAL person images with large variation in pose and scale. There are 14 annotated joint types (\ie forehead, neck, left/right shoulder, l/r elbow, l/r wrist, l/r waist, l/r knee and l/r ankle) and we combine the part labels into 6 semantic part types (\ie head, torso, upper arm, lower arm, upper leg and lower leg). We only use those images containing humans for training (1716 images) and validation (1817 images). We only experiment on this dataset because other datasets do not have both pose and part segment annotations.

\paragraph{Generation of joint proposals.} We apply the Faster R-CNN detector to produce human detection boxes, and perform a NMS procedure with detection score threshold $=0.6$ and box IOU overlap threshold = $0.6$. For each human detection box, we generate 6 joint proposals per joint type from the joint score map outputted by Pose FCN, using a NMS procedure with joint score threshold $= 0.2$ and proposal distance threshold $= 16$.

\paragraph{Generation of final pose configurations.} For each detection box, the FCRF selects and assembles joint proposals into a series of pose configurations, with pose score defined as the sum of all unary joint scores (in logarithm form). For each missing joint, we regard its unary score as 0.2. To combine pose configurations from all the detection boxes, we design a NMS prodedure which considers the overlap of head bounding box, upper-body bounding box, lower-body bounding box and whole-body bounding box inferred from the pose configurations. For two pose configurations, the one with a lower pose score will be filtered if their IOU overlap exceeds 0.65 for head boxes, or 0.5 for upper-body/lower-body boxes, or 0.4 for whole-body boxes.

\subsection{Human Pose Estimation}
Previous evaluation metrics (\eg PCK and PCP) do not penalize false positives that are not part of the groundtruth. So following~\cite{insafutdinov2016deepercut}, we compare our model with other state-of-the-arts by Mean Average Precision (mAP). Briefly speaking, pose configurations in $C^{pose}_I$ are first matched to groundtruth pose configurations according to the pose box overlap, and then the AP for each joint type is computed and reported. Each groundtruth can only be matched to one estimated pose configuration. Unassigned pose configurations in $C^{pose}_I$ are all treated as false positives.

\begin{table}[!tb]
%\vspace{-10pt}
\centering
\setlength{\tabcolsep}{5pt}
\resizebox{1\columnwidth}{!}{
\begin{tabular}{c | c c c c c c c c | c}
\toprule[0.2 em]
Method & Head & Shoulder & Elbow & Wrist & Hip & Knee & Ankle & U-Body & Total (mAP) \\ \midrule \midrule
Chen \& Yuille & 45.3 & 34.6 & 24.8 & 21.7 & 9.8 & 8.6 & 7.7 & 31.6 & 21.8 \\
Deeper-Cut & 41.5 & 39.3 & 34.0 & 27.5 & 16.3 & 21.3 & 20.6 & 35.5 & 28.6 \\
\hline
AOG-Simple & 56.8 & 29.6 & 14.9 & 11.9 & 6.6 & 7.3 & 8.6 & 28.3 & 19.4 \\
AOG-Seg & \textbf{58.5} & 33.7 & 17.6 & 13.4 & 7.3 & 8.3 & 9.2 & 30.8 & 21.2 \\
Our Model (w/o seg) & 56.8 & 52.1 & 42.7 & 36.7 & 21.9 & 30.5 & 30.4 & 47.1 & 38.7 \\
Our Model (final) & 58.0 & \textbf{52.1} & \textbf{43.1} & \textbf{37.2} & \textbf{22.1} & \textbf{30.8} & \textbf{31.1} & \textbf{47.6} & \textbf{39.2} \\
\bottomrule[0.1 em]
\end{tabular}
}
\caption{Mean Average Precision (mAP) of Human Pose Estimation on PASCAL-Person-Part.}
\label{table:pose_ap}
\end{table}

We compare our method with two other state-of-the-art approaches: (1) Chen \& Yuille~\cite{Chen_CVPR15}, a tree-structured model designed specifically for single-person estimation in presence of occlusion, using unary scores and image-dependent pairwise terms based on DCNN features; (2) Deeper-Cut~\cite{insafutdinov2016deepercut}, an integer linear programming model that jointly performs multi-person detection and multi-person pose estimation. These two methods both use strong graphical assembling models.
We also build two other baselines, which use simple And-Or graphs for assembling instead of the FCRF in our model. One is ``AOG-Simple'', which only uses geometric connectivity between neighbouring joints. The other one is ``AOG-Seg'', which adds part segment consistency features to ``AOG-Simple''. The part segment consistency features are the same as the segment-joint smoothness feature we use in the FCRF.
To test the effectiveness of our proposed part segment consistency, we also list the result of our model w/o the consistency features (``Our Model (w/o seg)''). The results are shown in Tab.~\ref{table:pose_ap}. Our model outperforms all the other methods, and by comparing our model with ``AOG-Simple'' and ``AOG-Seg'', we can see that a good assembling model is really necessary for challenging multi-person images like those in PASCAL.

\begin{table}[!b]
\centering
\setlength{\tabcolsep}{5pt}
\resizebox{1\columnwidth}{!}{
\begin{tabular}{c | c c c c c c c c | c}
\toprule[0.2 em]
Method & Forehead & Neck & Shoulder & Elbow & Wrist & Hip & Knee & Ankle & Ave. \\ \midrule \midrule
Chen \& Yuille & 37.5 & 29.7 & 51.6 & 65.9 & 72.0 & 70.5 & 79.9 & 78.6 & 60.7 \\
Deeper-Cut  & 32.1 & 30.9 & 37.5 & 44.6 & 53.5 & 53.9 & 65.8 & 67.8 & 48.3 \\
\hline
AOG-Simple & 33.0 & 33.2 & 66.7 & 82.3 & 90.5 & 89.7 & 101.3 & 101.1 & 74.7 \\
AOG-Seg & 32.2 & 31.6 & 59.8 & 72.4 & 85.1 & 85.7 & 97.1 & 92.7 & 69.6 \\
Our Model (w/o seg) & 27.7 & 26.9 & 33.1 & 40.2 & 47.3 & 51.8 & 54.6 & 53.4 & 41.9 \\
Our Model (final) & \textbf{26.9} & \textbf{26.1} & \textbf{32.7} & \textbf{39.5} & \textbf{45.3} & \textbf{50.9} & \textbf{52.3} & \textbf{51.8} & \textbf{40.7} \\
\bottomrule[0.1 em]
\end{tabular}
}
\caption{Average Distance of Keypoints (ADK) (\%) of Human Pose Estimation on PASCAL-Person-Part.}
\label{table:pose_adk}
\end{table}

Our proposed part segment consistency features not only help the overall pose estimation results, but also improve the accuracy of the detailed joint localization. Previous evaluation metrics (\eg PCP, PCK and mAP) treat any joint estimate within a certain distance of the groundtruth to be correct, and thus they do not encourage joint estimates to be as close as possible to the groundtruth. Therefore, we design a new evaluation metric called Average Distance of Keypoints (ADK). For each groundtruth pose configuration, we compute its reference scale to be half of the distance between the forehead and neck, then find the only pose configuration estimate among the generated pose configuration proposals that has the highest overlap with the groundtruth configuration. For each joint that is visible in both the groundtruth configuration and the estimated configuration, the relative distance (\wrt the reference scale) between the estimated location and the groundtruth location is computed. Finally, we compute the average distance for each joint type across all the testing images.

The result is shown in Tab.~\ref{table:pose_adk}. It can be seen that our model reduces the average distance of keypoints significantly for wrists and lower-body joints by employing consistency with semantic part segmentation.

\subsection{Human Semantic Part Segmentation}
We evaluate the part segmentation results in terms of mean pixel IOU (mIOU) following previous works~\cite{chen2014semantic,xia2015zoom}. In Tab.~\ref{table:seg}, we compare our model with two other state-of-the-art methods~\cite{chen2015attention,xia2015zoom} as well as one inferior baseline of our own model (\ie the output part label map $\ve{L}_s$ of the first-stage part FCN, without the help of pose information).

We also list the numbers of our model using the more advanced network architecture ResNet-101~\cite{chen2016deeplab} instead of VGG-16~\cite{chen2014semantic} for Part FCN. It can be seen that our model surpasses previous methods and the added pose information is effective for improving the segmentation results. When using ResNet-101, our model further boosts the performance to \textbf{64.39}\%.

\begin{table}[!tb]
\centering
\setlength{\tabcolsep}{5pt}
\resizebox{1\columnwidth}{!}{
\begin{tabular}{c | c c c c c c c | c}
\toprule[0.2 em]
Method & Head & Torso & U-arms & L-arms & U-legs & L-legs & Background & Ave. \\ \midrule \midrule
Attention~\cite{chen2015attention} & 81.47 & 59.06 & 44.15 & 42.50 & 38.28 & 35.62 & 93.65 & 56.39 \\
HAZN~\cite{xia2015zoom} & 80.76 & 60.50 & 45.65 & 43.11 & 41.21 & 37.74 & 93.78 & 57.54 \\
\hline
Our model (VGG-16, w/o pose) & 79.83 & 59.72 & 43.84 & 40.84 & 40.49 & 37.23 & 93.55 & 56.50 \\
Our model (VGG-16, final) & 80.21 & 61.36 & 47.53 & 43.94 & 41.77 & 38.00 & 93.64 & 58.06 \\
\hline
Our model (ResNet-101, w/o pose) & 84.95 & 67.21 & 52.81 & 51.37 & 46.27 & 41.03 & 94.96 & 62.66 \\
Our model (ResNet-101, final) & \textbf{85.50} & \textbf{67.87} & \textbf{54.72} & \textbf{54.30} & \textbf{48.25} & \textbf{44.76} & \textbf{95.32} & \textbf{64.39} \\
\bottomrule[0.1 em]
\end{tabular}
}
\caption{Mean Pixel IOU (mIOU) (\%) of Human Semantic Part Segmentation on PASCAL-Person-Part.}
\label{table:seg}
\end{table}

\begin{table}[!b]
\centering
\setlength{\tabcolsep}{5pt}
\resizebox{1\columnwidth}{!}{
\begin{tabular}{c | c c c c c c c | c}
\toprule[0.2 em]
Method & Size XS & Size S & Size M & Size L \\ \midrule \midrule
Attention~\cite{chen2015attention} & 37.6 & 49.8 & 55.1 & 55.5 \\
HAZN~\cite{xia2015zoom} & 47.1 & 55.3 & 56.8 & 56.0 \\
\hline
Our model (ResNet-101, w/o pose) & 40.4 & 54.4 & 60.5 & 62.1 \\
Our model (ResNet-101, final) & \textbf{53.4} & \textbf{60.9} & \textbf{63.0} & \textbf{62.8} \\
\bottomrule[0.1 em]
\end{tabular}
}
\caption{Mean Pixel IOU (mIOU) (\%) of Human Semantic Part Segmentation \wrt Size of Human Instance on PASCAL-Person-Part.}
\label{table:seg_scalewise}
\end{table}

Besides, we evaluate part segmentation \wrt different sizes of human instances in Tab.~\ref{table:seg_scalewise}, following~\cite{xia2015zoom}.
Our model performs especially well for small-scale people, surpassing other state-of-the-arts by over \textbf{5}\%.
\begin{figure*}[!tb]
\begin{center}
   \includegraphics[width=0.82\textwidth]{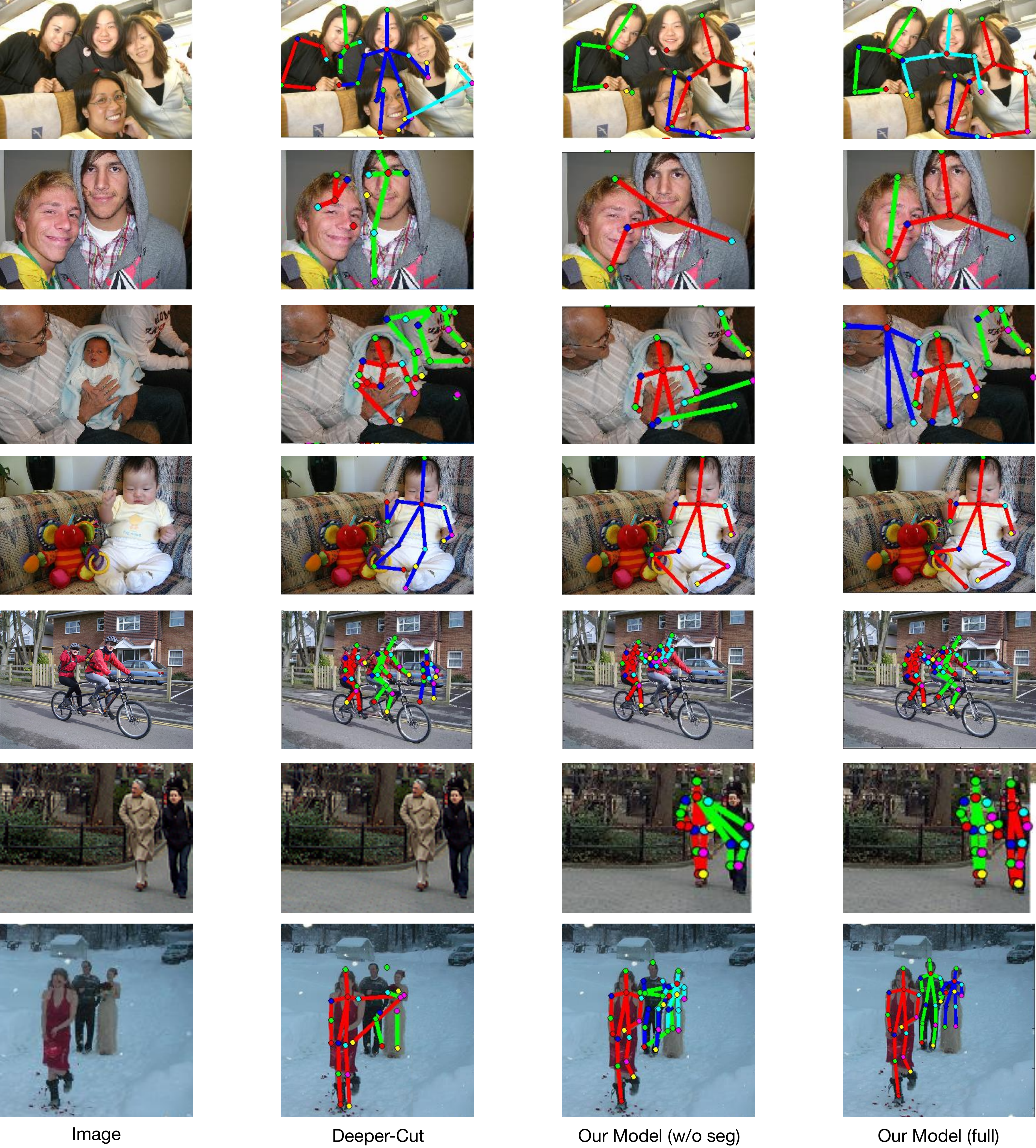}
\end{center}
\caption{Visual comparison of human pose estimation on PASCAL-Person-Part~\cite{chen2014detect}. Our full model is compared against Deeper-Cut~\cite{insafutdinov2016deepercut} and a variant of our model (``Our Model (w/o seg)'') that doesn't consider part segment consistency.}
\label{fig:vis_pose}
%\vspace{-10pt}
\end{figure*}
\begin{figure*}[!tb]
\begin{center}
   \includegraphics[width=1.0\textwidth]{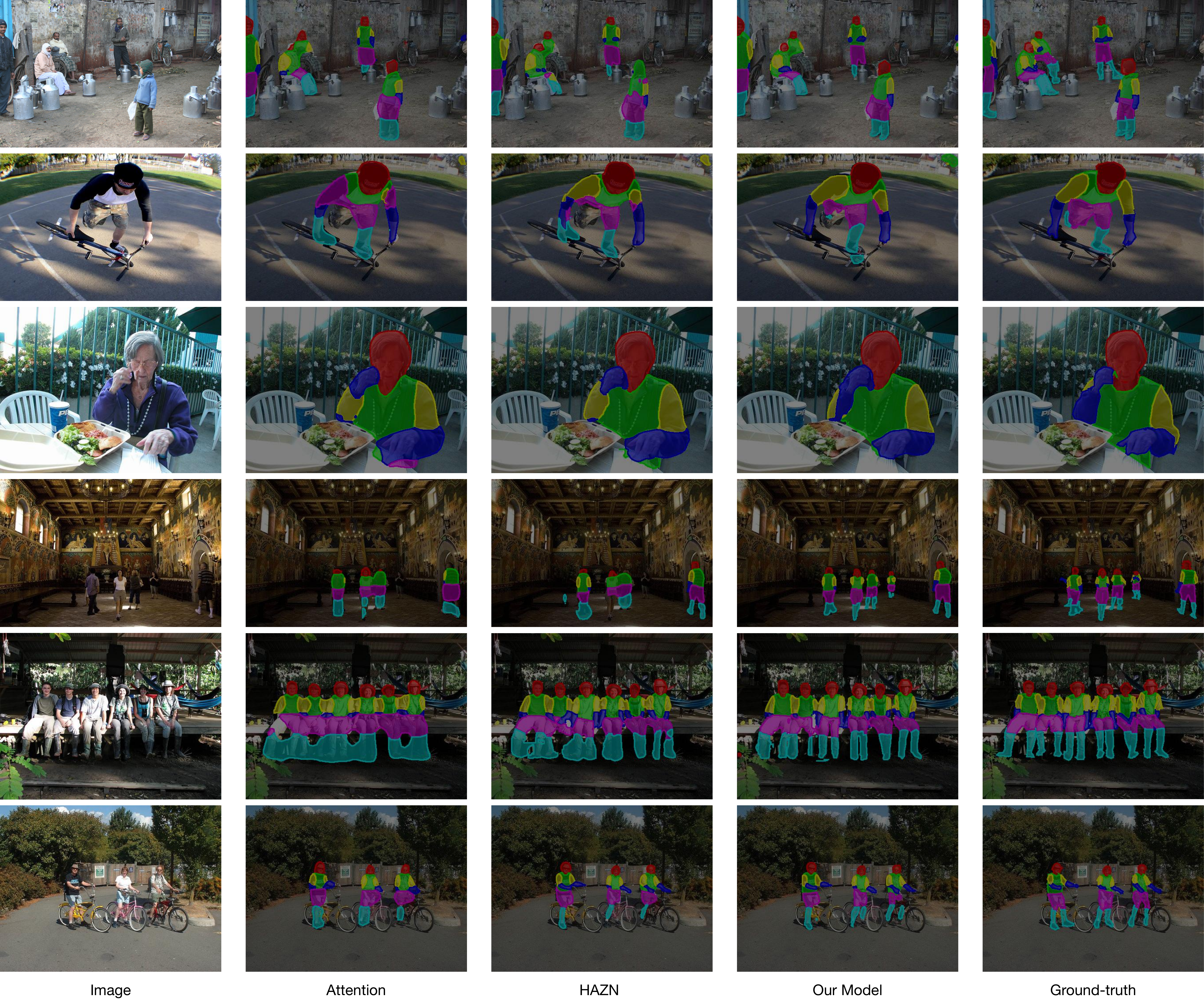}
\end{center}
\caption{Visual comparison of human semantic part segmentation on PASCAL-Person-Part~\cite{chen2014detect}. Our method is compared against two recent state-of-the-art methods: Attention~\cite{chen2015attention} and HAZN~\cite{xia2015zoom}.}
\label{fig:vis_seg}
\end{figure*}

\subsection{Qualitative Evaluation}
\paragraph{Human pose estimation.} In Fig.~\ref{fig:vis_pose}, we visually demonstrate our pose estimation results on PASCAL-Person-Part, comparing them with the recent state-of-the-art Deeper-Cut~\cite{insafutdinov2016deepercut} and also a sub-model of ours (``Our Model (w/o seg)'') which does not consider part segment consistency. This shows that our model gives more accurate prediction of heads, arms and legs, and is especially better at handling people of small scale (see the $6_{th}$ and $7_{th}$ row of Fig.~\ref{fig:vis_pose}) and extra large scale (see the first two rows of Fig.~\ref{fig:vis_pose}).

\paragraph{Human semantic part segmentation.} Fig.~\ref{fig:vis_seg} visually illustrates the advantages of our model over two other recent methods, Attention~\cite{chen2015attention} and HAZN~\cite{xia2015zoom}, which adopt the same basic network structure as ours. Our model estimates the overall part configuration more accurately. For example, in the $2_{rd}$ row of Fig.~\ref{fig:vis_seg}, we correctly labels the right arm of the person while the other two baseline methods label it as upper-leg and lower-leg. Furthermore, our model gives clearer details of arms and legs (see the last three rows of Fig.~\ref{fig:vis_seg}), especially for small-scale people.

\section{Conclusion}
In this paper, we demonstrate the complementary properties of human pose estimation and semantic part segmentation in complex multi-person images. We present an efficient framework that performs the two tasks iteratively and improves the results of each task. For human pose estimation, we adopt a fully-connected CRF that jointly performs human instance clustering and joint labeling, using deep-learned features and part segment based consistency features. This model gives better localization of joints, especially for arms and legs. For human semantic segmentation, we train a FCN that uses estimated pose configurations as shape and location priors, successfully correcting local confusions of people and giving clearer details of arms and legs.

We also adopt an effective ``auto-zoom'' strategy that deals with object scale variation for both tasks and helps reduces the inference time of the CRF by a factor of 40. We test our approach on the challenging PASCAL-Person-Part dataset and show that it outperforms state-of-the-art methods for both tasks.

\section{Acknowledgements}
We are deeply grateful for the support from ONR N00014-15-1-2356, NSF award CCF-1317376 and Army Research Office ARO 62250-CS, and also for the free GPUs provided by NVIDIA.

\newpage
{\small
\bibliographystyle{ieee}
\bibliography{cvpr17}
}
\end{document}